\documentclass[11pt]{article}

\usepackage[whole]{bxcjkjatype}
\usepackage{amsmath,bm,amssymb}

\usepackage[final]{acl}

\usepackage{times}
\usepackage{latexsym}

\usepackage[T1]{fontenc}
\usepackage[utf8]{inputenc}

\usepackage{microtype}

\usepackage{inconsolata}

\usepackage{graphicx}
\usepackage{multirow}

\title{Timesteps of Mamba Align with Human Reading Times}

\author{
 \textbf{Yuji Yamamoto\textsuperscript{1,2*}}\ 
 \textbf{Shinnosuke Isono\textsuperscript{2*}}\ 
 \textbf{Yoshinobu Kawahara\textsuperscript{3,5}}\ 
 \textbf{Sho Yokoi\textsuperscript{2,4,5}}
\\
\\
 \textsuperscript{1}SOKENDAI\ 
 \textsuperscript{2}NINJAL\ 
 \textsuperscript{3}The University of Osaka\ 
 \textsuperscript{4}Tohoku University\ 
 \textsuperscript{5}RIKEN
\\
 \texttt{\{yuji.yamamoto, s-isono, yokoi\}@ninjal.ac.jp \ kawahara@ist.osaka-u.ac.jp}
}

\begin{document}
\maketitle
\def\thefootnote{*}\footnotetext{Equal contribution}\def\thefootnote{\arabic{footnote}}

% 本研究では，状態空間言語モデル Mamba の暗黙的な「単語の処理時間」と，人間が文章を読む際の各単語の読む時間の間に強い関係があることを，経験・理論の両面から述べる．
% Mamba は post-Transformer の候補のひとつとなっている再帰的な言語モデルで，とくに，各時刻（各単語）$t$に対して，離散化ステップと呼ばれる状態遷移時間$\Delta_t$がモデルの各層において入力依存に割り振られる．
% 我々は，この Mamba の離散化ステップが，ヒトの読み時間を予測する変数としてTransformer言語モデルのサプライザルという強力なベースラインを経験的に匹敵すること，また各層の離散化ステップがそれぞれ異なる言語的な特徴に反応することを示す．
% さらに，読み時間モデリングに特に寄与する層では，モデルの構造上「長期的な情報の保持」を行うのに有効であり，人工データを用いた実験を通してこれが確かに記憶と想起のためのモジュールとして機能していることを確認する．
% 状態空間言語モデルが，工学的に面白いモデルであるばかりでなく，認知モデリングのための道具としても注目に値することを示したい．
\begin{abstract}
This study demonstrates an alignment of per-word processing time in a popular state-space language model Mamba and human readers. In Mamba, the recurrent state transition at each layer conceptually takes some duration of time, the discretization timestep $\Delta_t$, determined dynamically in response to the input. Using a naturalistic reading dataset, we show that the per-word timestep from Mamba is a significant predictor of human reading times, and remains significant even when known predictors such as GPT-2 surprisal are controlled for. We further suggest, through formal analysis of Mamba's architecture and internal dynamics, that Mamba can serve as a new, valuable lens to look at human real-time language processing with ever-updated memory, because it allows us to look at how each module (layer) weighs short- and long-term information retention, and how noise may interact with dynamic, continuous memory representation. Code is available online.\footnote{Code available at \url{https://osf.io/vnw5e/overview?view_only=93ad704fc6ea44438f3d3538b4b682eb}}
\end{abstract}

\section{Introduction}
Human language inherently unfolds over time. In the rise of fluent artificial large language models (LMs), it has become interesting to ask how time may be represented in (unidirectional variants of) them, and how their time is related to temporal dynamics of language processing in humans.

\begin{figure}[t]
  \centering
  \includegraphics[width=\linewidth]{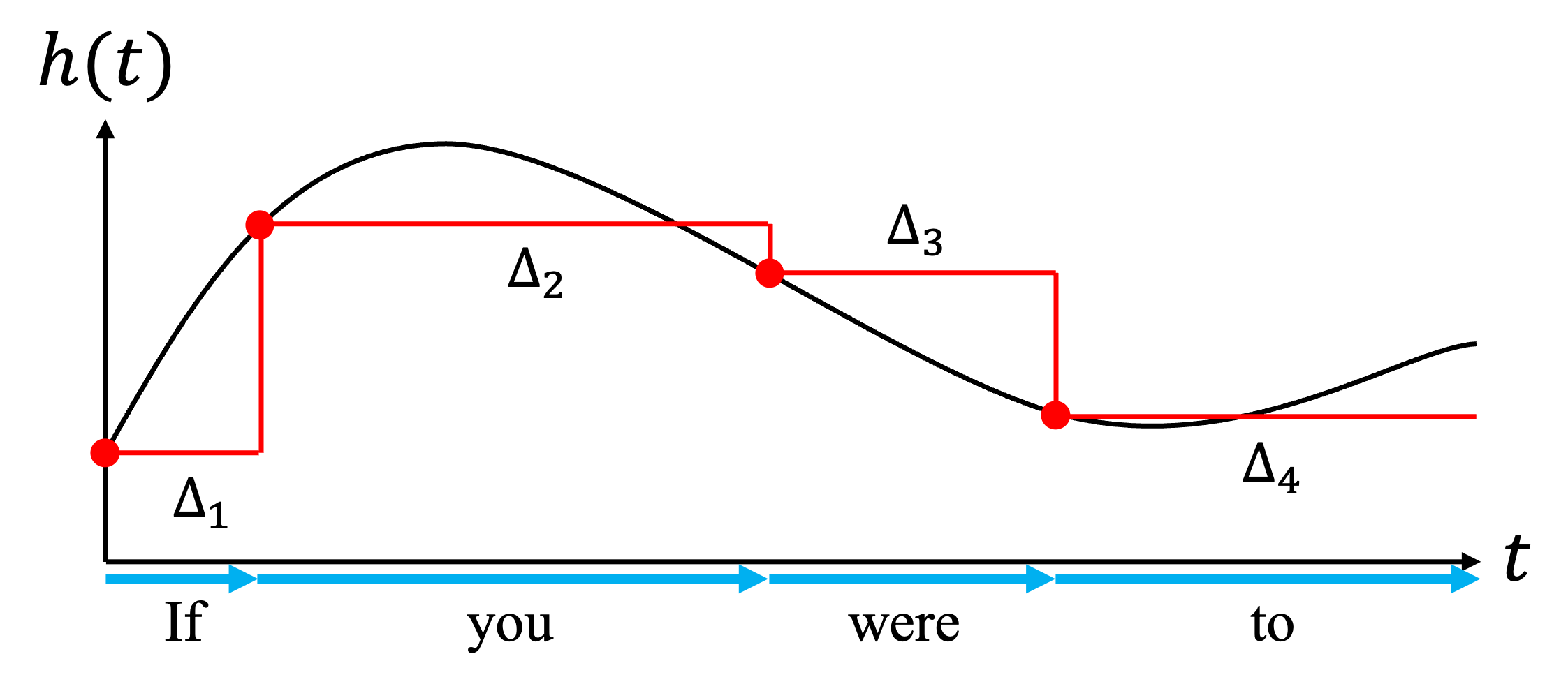}
  \caption{Illustration of discretization of a continuous-time state transition $h(t)$ using the input-dependent timestep~$\Delta_t$.}
  % Mamba discretizes continuous-time function using timesteps that are input-dependent rather than fixed.
  \label{fig:image}
\end{figure}

\emph{Mamba} \cite{gu2024mamba}, a popular architecture among LMs based on a state-space model (SSM), is interesting in this respect, since it has a notion of internal timesteps, which modulates state transitions in response to input. SSMs represent sequential data as continuous state transitions and are characterized by their ability to handle long-range dependencies in a stable manner.
Among them, Mamba is designed as a selective SSM that dynamically modulates state transitions by changing the size of timesteps in response to the input, offering greater flexibility than other SSMs that assume fixed transition dynamics.
As illustrated in Figure \ref{fig:image}, this property enables the model to immediately reflect changes of context and importance of particular input in its internal states, allowing adaptive representation learning even for data with complex and non-stationary structures.

\begin{figure}[t]
  \centering
  \includegraphics[width=\linewidth]{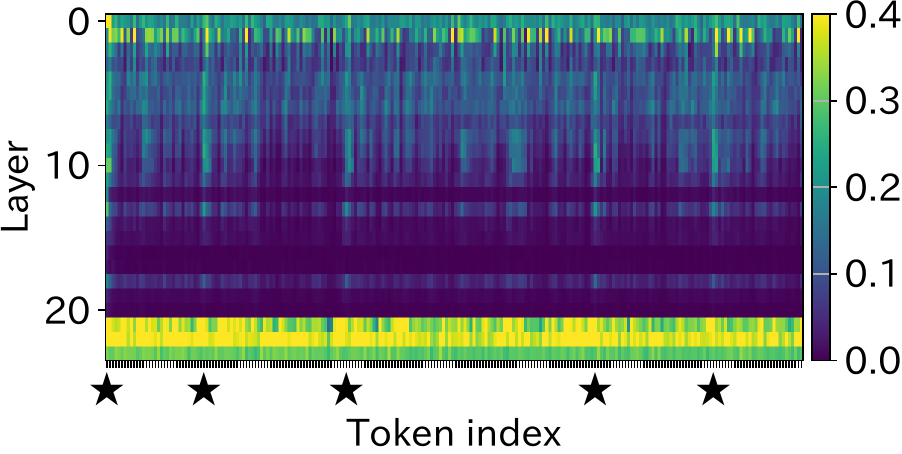}
  \caption{Discretization timesteps $\bar\Delta_t$ of each of the 24 layers of Mamba-130M, in response to the first five sentences from the Natural Stories Corpus.
  ``★'' indicates the beginning of a sentence.
  }
  %第4層から第20層までは文境界付近でピークを取る層が複数見られるが，第12,16,17層は文境界でピークを取らないことが分かる．}
  \label{fig:delta_t_vis}
\end{figure}

In this study, we demonstrate a word-level alignment between Mamba's internal timesteps and human processing time. After establishing that Mamba's internal timesteps can be seen as processing time (\S\ref{sec:processing-time}), we show that the sizes of the timesteps that Mamba assigns to each word in a naturalistic corpus is predictive of the time that human readers spend on each word (\S\ref{sec:experiment}). The performance of the best-performing layer is comparable to that of GPT-2 surprisal, a variable known to be a powerful predictor of human reading times, and remains significant even when GPT-2 surprisal and other known predictors are controlled for.
We also conducted formal analyses looking into how Mamba's architecture and internal dynamics process language over time, and argue that the model can serve as a new, valuable lens to look into human real-time language processing with ever-updated memory (\S\ref{sec:lens}). Specifically, we observe that the model's transition matrix eigenspectrum specifies a different degree of memory retention ability for each layer, enabling the entire model to process both short- and long-term information. We also theoretically show that the size of the timesteps corresponds to uncertainty in state transition, and suggest that this offers a new insight into noisy language processing.
% We theoretically show that the magnitude of the discretization timestep $\Delta_t$ corresponds to uncertainty in state transitions (\S\ref{sec:uncertainty}). 
% We then empirically conduct reading-time modeling using the discretization timestep $\Delta_t$ to examine whether Mamba's implicit processing time corresponds to human reading time (\S\ref{sec:experiment}).
% The results revealed the following points:
% \begin{itemize}
  % \item Timesteps from a majority of layers are predictive of the human reading time. The performance of the best-performing layer is comparable to that of GPT-2 surprisal, a variable known as a powerful predictor.
  % \item Layer
  % \item Some layers' timesteps $\Delta_t$ retain significant predictive power even when well-known variables such as GPT-2 surprisal are added to the baseline.
  % \item Timesteps $\Delta_t$ for word $w_t$ from different layers correlate with different linguistic features, suggesting a distinct role of each layer (\S\ref{sec:linguistic_analysis}).
% \end{itemize}

% In addition, timesteps $\Delta_t$ for a word $w_t$ from different layers correlate with distinct linguistic features, suggesting distinct roles of each layer (\S\ref{sec:linguistic_analysis}). 
% Both mathematical analysis and empirical experiments using synthetic data show that the layers contributing most to reading-time modeling are effective at retaining information over the long term (\S\ref{sec:passkey},\S\ref{sec:eigen}).

\section{Preliminaries}

\subsection{Mamba}
\label{sec:mamba}

% In this subsection, 
We provide an overview of Mamba; for further details, see \cite{gu2024mamba}. 

Mamba is a popular state-space model (SSM)-based language model. 
Unlike Transformers, SSMs do not process the entire sequence simultaneously but instead handle input and output recursively and sequentially on a word-by-word basis, in a similar fashion to RNNs and human readers.
Because SSMs are recursive, a naive implementation would require computation time proportional to the input sequence length. 
However, since the hidden-state update equations of SSMs contain no nonlinearities, the model can be accelerated using parallel scan algorithms \cite{blelloch1990prefix}, making large-scale language model training feasible.
Mamba, one of the representative state-space models, as well as Linear Attention \cite{yang2024fla} models that also employ linear state updates, are employed in the development of highly efficient large language models \cite{zuo2024falcon,networks2025plamo,team2025kimi,blakeman2025nvidia}.

Mamba has multiple layers
(for example, the Mamba-130M model used in the current experiment consists of 24 layers\footnote{{https://huggingface.co/state-spaces/mamba-130m-hf}}).
Each layer incorporates a discretized form of a state-space model. 
While state-space models are defined as differential equations in continuous time, they are approximated as difference equations in discrete time when applied to real data, particularly discrete data such as language (Figure \ref{fig:image}). 
In standard discretization procedures, a constant discretization timestep $\Delta$ is used regardless of the time index $t$; 
in contrast, Mamba assigns an \emph{input-dependent discretization timestep} $\Delta_t$ to each timestep $t$.

% Δt は mamba の記憶と忘却を司るパラメータだ
The discretization timestep $\Delta_t$ acts as a gate in each layer of Mamba, governing memory retention and forgetting.
Below, we formally explain this by referring to the specific update equations of Mamba.
Given an input $\bm{x}_t \in \mathbb{R}^d$, Mamba applies the following equations independently to each element $x_{ti}$ of $\bm{x}_t$ in parallel $d$ times, and outputs the vector $\bm{y}_t\in\mathbb{R}^d$ obtained by collecting the resulting outputs $y_{ti}$.
Mamba consists of two equations: one updates a component called the hidden state $\bm{h}_t\in\mathbb{R}^n$, which corresponds to memory, using the input $x_{ti}$, and the other generates the output $y_{ti}$ from the hidden state~$\bm{h}_t$:
% \footnote{In Mamba, the SSM is applied in parallel to each dimension $i$ of the input $\bm{x}_t$. That is, if the hidden representation $\bm{x}_t$ is a $d$-dimensional vector, each layer contains $d$ SSMs operating in parallel.}
\begin{align}
  \bm{h}_t &= \overline{A}_t \bm{h}_{t-1} + \overline{B}_t x_{ti} \label{eq:hidden-states} \\
  y_{ti} &= C_t \bm{h}_t + D x_{ti}.
\end{align}
Here, the state transition matrix $\overline{A}_t \in\mathbb{R}^{n\times n}$ and input matrix $\overline{B}_t\in\mathbb{R}^{n\times 1}$, which represent the weights for memory and input in the state update equation, are defined as follows with an input-dependent discretization timestep~$\Delta_t~\in~(0,~\infty)$:
% specified in Equation \ref{}:
\begin{align}
  \overline{A}_t &= \exp(\Delta_t A) ~\left(= \exp(A)^{\Delta_t}\right) \label{eq:ssm-param1} \\
  \overline{B}_t &= \Delta_t W_B \bm{x}_t \label{eq:ssm-param2}\\
  \Delta_t &=  \log(1 + \exp(W_\Delta\bm{x}_t+\bm{b}_\Delta)). \label{eq:delta}
\end{align}
As can be seen from the above equations, the discretization timestep $\Delta_t$ plays an important role in modulating the weight coefficients for memory and input. 
For example, when $\Delta_t$ is large, the elements of the matrix $\overline{A}_t$ approach zero,
\footnote{In Mamba, the transition matrix $\exp(A)$ is implemented to be diagonal, with elements constrained to lie within the range $(0, 1]$.}
thereby reducing the influence of the previous memory $\bm{h}_{t-1}$.
At the same time, the absolute values of the input weight coefficients $\overline{B}_t$ increase, strengthening the influence of the input $\bm{x}_t$.\footnote{%
Conversely, when $\Delta_t$ is small, that is, close to zero, the elements of $\overline{A}_t$ approach one and those of $\overline{B}_t$ approach zero, preserving the previous memory $\bm{h}_{t-1}$ while discarding the input $\bm{x}_t$.
}

\subsection{Reading Time Modeling}

Reading time modeling is a research approach that aims to quantitatively identify factors contributing to cognitive load during sentence comprehension by explaining variations in the reading time that readers spend reading each word in a text.
Reading time is considered to reflect comprehension difficulties and information processing complexity, and psycholinguists have attempted to identify factors contributing to cognitive load by statistically analyzing its variation.
Many previous studies in this approach use naturalistic reading time datasets, which are collections of reading time data from tens to hundreds of participants reading naturalistic texts \cite[e.g.][]{demberg-keller-2008-data, wilcox-etal-2023-testing}.
The data is modeled by some regression models that contain the variable of theoretical interest plus control variables that are known to affect reading time, such as word length, lexical frequency, syntactic complexity, and contextual predictability.
The significance of an explanatory variable suggests that the variable correlates with language processing difficulty, leading to a constructivist understanding of which information is crucial for reading comprehension.
% \cite{kuribayashi2022workingmemory}.
% \textcolor{red}{こっちに置き換え？文章参考にしてないけど https://aclanthology.org/2022.emnlp-main.712}

Most studies interested in the relation between large language models and human reading today use the next word probability to link the models and reading times. The \emph{surprisal theory} \cite{hale-2001-probabilistic, levy-2008-expectation} predicts that the processing cost of word $w$ given the preceding context $C$ is proportional to $-\log p(w\mid C)$. The combination of the Transformer architecture and the surprisal theory has proven successful in reading time modeling for naturalistic texts in many languages \cite{wilcox-etal-2023-testing}. Studies have also found, however, that models that are better from an engineering perspective, with more parameters, lower perplexity, or larger context window, are not necessarily better at predicting human reading times than smaller models \cite{oh-schuler-2023-surprisal,kuribayashi-etal-2022-context,kuribayashi-etal-2021-lower}. While these studies seek a link between models and humans at the computational level in the sense of Marr \cite{marr-1982-vision}, other studies seek to link Transformers and humans at the algorithmic level, e.g., comparing the attention mechanism to the memory retrieval mechanism in humans \cite{ryu-lewis-2021-accounting,yoshida-etal-2025-attention}. 
% 大規模言語モデルを用いた読み時間モデリングでは，モデルが単語に付与する確率が主に用いられてきた．ある単語$w$の文脈$C$での処理負荷が$-\log p(w\mid C)$に比例するというサプライザル理論\cite{hale-2001-probabilistic, levy-2008-expectation}と組み合わせることで読み時間が予測される．Transformerに基づくサプライザルが読み時間を予測することが多くの言語で確かめられている一方\cite{wilcox-etal-2023-testing}，モデルサイズやコンテクスト幅が大きいとむしろヒトの傾向から離れていくことも指摘されている\cite{oh-schuler-2023-surprisal,kuribayashi-etal-2022-context}．アルゴリズム・レベルでは，Transformerのアテンションが，心理言語学で仮定されてきたヒトの記憶想起（retrieval）と類似すると指摘されている\cite{ryu-lewis-2021-accounting,yoshida-etal-2025-attention}．ただ，Transformerのアテンションは各トークンについての情報に直接アクセスするのに対し，ヒトの記憶では情報は入力の時系列に沿って重ね合わせて保持されているとみられている\cite{oberauer-etal-2012-modeling}．この点で，入力に応じて記憶の状態が常に変化していくSSMが，Transformerとは異なった形でヒトの処理の特徴を捉えている可能性がある．

The current study explores yet another approach to reading time modeling, with a different type of model (Mamba) and a different linking hypothesis (intepreting $\Delta_t$ as processing time) than the standard combination of Transformer and surprisal.

\section{Discretization Timesteps of Mamba as Word Processing Time}\label{sec:processing-time}
% <2.1 の冒頭文のふたつめ>
% また、このあと説明する離散化時間Δtという要素が、単語tの「処理時間」に対応していることと、これが入力に応じて変化することを説明する。
We describe that the discretization timestep $\Delta_t$ corresponds to the ``processing time'' of word $t$ and that this quantity adapts to the input.

% <ぱらぐらふひとつめ; 対応する話> ← 主題
% Notably, the discretization timestep $\Delta_t$ は、単語の「処理時間」に対応するものである。
% <intro から2¶をくっつけて移動>
Notably, the input-dependent discretization timestep $\Delta_t$ (Eq. \eqref{eq:delta}) in the language model Mamba can be interpreted as the implicit ``processing time'' involved in state transitions in continuous time.
Typically, in the discretization of differential equations, the time evolution in continuous time is approximated using a predetermined fixed step size.
In contrast, in Mamba, when modeling continuous-time state transitions as discrete recurrence relations, the time is discretized into timesteps whose length varies depending on the input to each token (Figure \ref{fig:image} and \ref{fig:delta_t_vis}).
In other words, how long it takes to update the state is controlled by the input tokens．

% <ふたつめ; Δt は input-dependent で、なので人の読み時間っぽいな〜>
Mamba potentially captures the processing time allocated to each token through the input-dependent discretization timestep $\Delta_t$.
Mamba does not process each token on a fixed time scale; instead, it dynamically adjusts the timesteps $\Delta_t$ according to the nature of the input tokens.
For example, when the timestep $\Delta_t$ is larger, the model continues its time evolution over a longer period while holding the same state.
This can be interpreted as a mechanism that adjusts the processing time based on the input.
This property is functionally analogous to reading time, where humans pause longer on words or contexts that require a higher cognitive load during comprehension.
% From this perspective, Mamba could potentially acquire context-appropriate processing time for each word without explicit supervision through the input-dependent discretization timestep~$\Delta_t$.

\section{Mamba's Timesteps are Predictive of Human Reading Times}
\label{sec:experiment}
% \section{実験}
The current experiment investigates whether Mamba's ``processing time'', $\Delta_t$, is predictive of human word-by-word reading times, and whether its predictive power is beyond that of other known variables.
% 本節では，Mambaの``処理時間''である$\Delta_t$と人間の読み時間との間に統計的に有意な関係があるかを見る．

% 分析では，次の点を問うた．
% \begin{itemize}
%     \item 各層の$\Delta_t$はそれぞれ，読み時間を予測するか？
%     \item 各層の$\Delta_t$を予測力が高くなるように組み合わせたとき，その予測力はGPT-2サプライザルと比べて高いか？
%     \item $\Delta_t$の予測力は，読み時間を予測する既知の基本的な変数を含めても残るか？
% \end{itemize}

\subsection{Settings}
% \subsection{実験設定}
%実験にはNatural Storiesコーパス\cite{futrell2021natural}を用いた．同コーパスは，読み時間モデリングに適するよう，人為的に複雑な構文を交えつつ，全体として自然に読めるように編集された10個の物語（英語，485文，10,245単語）からなる．これに，母語話者181人が自己ペース読文法\cite{just-etal-1982-paradigms}で行った単語ごとの読み時間が付されている．自己ペース読文法では，一度に一単語のみが提示され，読み手は指定のボタンを押すことで自分のペースで次の単語に進む．読み戻りはできない．各単語が表示されている時間を読み時間として扱う．本分析では，事前に参加者ごとの切片のみの線形混合モデル$\log(RT)\sim 1+(1\mid\mathit{participant})$をフィットし，コーパス中の単語ごとに残差の平均値をとって従属変数とした．以下，この従属変数を単に読み時間と呼ぶ．
%\Delta_t$の値はMamba-130m\footnote{\url{https://huggingface.co/state-spaces/mamba-130m-hf}}からとり，各時点での $\bm{\Delta}_t \in\mathbb{R}^{d}$ は次元方向に平均を取り，subword に分かれた $\Delta_t$ は最大値を取ることで単語レベルに集約した．
\subsubsection{Datasets}
We used two reading time datasets.

\paragraph{Natural Stories} Natural Stories \cite{futrell2021natural} consists of data from 181 native speakers of English reading 10 stories (10,245 words in total) using the self-paced reading method \cite{just-etal-1982-paradigms}. In this method, one word is presented at a time, and participants press a key to proceed to the next word; no backtracking is allowed. The reading time for a word refers to the duration for which the word is displayed.

\paragraph{OneStop} OneStop \cite{berzak-etal-2025-onestop}  (the ``ordinary reading'' subset) consists of reading time data from 180 native speakers of English reading 30 articles using the eye-tracking-during-reading method. The articles come with the original ``advanced'' version (19,428 words) and the simplified ``elementary'' version (15,737 words); both were used in the current analysis. We analyze first-pass time (the time between the entry to the region and the first exit to the left or right in the first pass) and regression-path duration (the time between the first entry to the region and the first exit to the right), since they can be seen as conceptual counterparts of Mamba's $\Delta_t$: they reflect the processing time before any information from the right is accessed.

% The $\Delta_t$ values were taken from Mamba-130M and Mamba-2.8B. $\bm{\Delta}_t\in\mathbb{R}^{d}$ for each $t$ was averaged across the dimensions $d$, and was further aggregated at the word-level by taking the maximal value, when the word was analyzed into multiple subwords.

\subsubsection{Predictors}
\paragraph{Discretization timestep}
The discretization timestep $\Delta_t$ was taken from Mamba-130M and Mamba-2.8B.
As described in \S\ref{sec:mamba}, the update equation \eqref{eq:hidden-states} in Mamba is applied in parallel across $d$ dimensions, so the discretization timestep is obtained as a $d$-dimensional vector $\bm{\Delta}_t$.
In the reading-time modeling analysis, we take the average $\bar{\Delta}_t = \sum_i (\bm{\Delta}_t)_i$ to aggregate the discretization timestep into one scalar for each layer and each timestep, and it was further aggregated at the word-level by taking the maximal value, when the word was analyzed into multiple subwords.

\paragraph{Control predictors} Besides $\Delta_t$, we included the following as control variables: the number of characters in the word, corpus frequency of the word, position of the word in the sentence in the story, Mamba surprisal, and GPT-2 surprisal.\footnotemark{}

\footnotetext{The GPT-2 surprisal values were calculated using the code provided by \cite{oh-schuler-2023-surprisal}.}

\subsubsection{Regression analysis}
The three reading time metrics (one from Natural Stories and two from OneStop) were modeled separately. The dependent measure was the by-token reading time, obtained by taking the mean residual reading times from a linear mixed model $\log(RT)\sim 1+(1\mid\mathit{participant})$.

In each analysis, linear regression models were fitted to the reading times and evaluated by 10-fold cross-validation repeated 50 times. The predictive power of the models was quantified by their per-word mean squared error (MSE) of their prediction for the unseen data. The predictive power of a particular variable of interest was evaluated by comparing the MSEs of two models with and without that variable using the permutation test with $\alpha=0.05$. Since we test each layer separately, $p$-values were adjusted for multiple comparisons by the Holm method. Linear models predicting $RT_t$ at word $w_t$ included variables associated not only with $w_t$ but also with $w_{t-2}$ and $w_{t-1}$ to cover spillover effects, i.e., delayed reading slowdown effects due to the cost of preceding words, observed in many reading studies \cite{wilcox-etal-2023-testing}.
%いずれの分析も，読み時間を予測する線形回帰モデルを構築し，10-fold交差検証で予測誤差（平均二乗誤差，MSE）を評価することで行った．特定の変数をモデルに含めることによるMSEの減少（$\Delta$MSE）が0より有意に大きいかを，置換検定により検定した．有意水準は$\alpha=0.05$とし，層別の評価の場合には，Holm法による$p$値の補正を行った．スピルオーバー（ある単語に起因する処理負荷が次単語以降の読み時間にも反映されること）を考慮し，$RT_t$の予測には$w_{t-2},w_{t-1},w_t$に対応する従属変数を用いた\cite{wilcox-etal-2023-testing}．

 We considered four kinds of regression models, which ask different questions:
\begin{itemize}
    \item Without any control variables: Does $\bar\Delta_t$ align with human reading times?
    \item With low-level (i.e., non-surprisal) variables: Does $\bar\Delta_t$ explain variance not explained by low-level features of the linguistic stimuli?
    \item With low-level variables and Mamba surprisal: Does $\bar\Delta_t$ explain variance even when predictability effects \cite{hale-2001-probabilistic,levy-2008-expectation} are taken into account?
    \item With low-level variables and GPT-2 surprisal: Does $\bar\Delta_t$ explain variance even when the state-of-the-art predictor of predictability effects \cite{oh-schuler-2023-surprisal} is taken into account?
\end{itemize}

We also asked whether these layers predict reading times independently. We searched for the combination of layers with the best predictive performance. Starting from the layer with the highest $\Delta$MSE in the layer-wise evaluation, we considered one model at a time for the inclusion to the regression model, and included only when its inclusion resulted in a significant improvement of $\Delta$MSE.

\begin{table*}[t]
    \centering
    \small
    \begin{tabular}{r|rrr|rrr|rrr|rrr}
        \hline
        & & & & & & & \multicolumn{3}{c|}{Low-level +} & \multicolumn{3}{c}{Low-level +}\\
        Baseline$\to$ & \multicolumn{3}{c|}{Intercept-only} & \multicolumn{3}{c|}{Low-level variables} & \multicolumn{3}{c|}{Mamba surprisal} & \multicolumn{3}{c}{GPT-2 surprisal}\\
        $\downarrow$Layer & $\Delta$MSE & $R^2$ & $\beta$ & $\Delta$MSE & $R^2$ & $\beta$ & $\Delta$MSE & $R^2$ & $\beta$ & $\Delta$MSE & $R^2$ & $\beta$\\
        \hline
        0 & $0.01^{\phantom{***}}$ & $0.00$ &  & $0.05^{***}$ & $0.49$ & ${}^{++-}$ & $0.07^{***}$ & $0.53$ & ${}^{++-}$ & $0.06^{***}$ & $0.54$ & ${}^{++-}$\\
1 & $0.06^{**\phantom{*}}$ & $0.01$ & ${}^{+++}$ & $0.02^{*\phantom{**}}$ & $0.49$ & ${}^{++-}$ & $0.03^{***}$ & $0.53$ & ${}^{++-}$ & $0.03^{***}$ & $0.53$ & ${}^{++-}$\\
2 & $0.05^{**\phantom{*}}$ & $0.01$ & ${}^{+++}$ & $0.02^{*\phantom{**}}$ & $0.49$ & ${}^{++-}$ & $0.03^{***}$ & $0.53$ & ${}^{++-}$ & $0.03^{**\phantom{*}}$ & $0.53$ & ${}^{++-}$\\
3 & $0.00^{\phantom{***}}$ & $0.00$ &  & $0.02^{*\phantom{**}}$ & $0.49$ & ${}^{++-}$ & $0.01^{\phantom{***}}$ & $0.53$ &  & $0.01^{\phantom{***}}$ & $0.53$ & \\
4 & $0.00^{\phantom{***}}$ & $0.00$ &  & $0.00^{\phantom{***}}$ & $0.49$ &  & $0.00^{\phantom{***}}$ & $0.52$ &  & $0.00^{\phantom{***}}$ & $0.53$ & \\
5 & $0.02^{\phantom{***}}$ & $0.00$ &  & $0.01^{\phantom{***}}$ & $0.49$ &  & $0.01^{\phantom{***}}$ & $0.53$ &  & $0.01^{\phantom{***}}$ & $0.53$ & \\
6 & $0.07^{***}$ & $0.01$ & ${}^{++-}$ & $0.01^{\phantom{***}}$ & $0.49$ &  & $0.00^{\phantom{***}}$ & $0.52$ &  & $0.00^{\phantom{***}}$ & $0.53$ & \\
7 & $0.16^{***}$ & $0.02$ & ${}^{+++}$ & $0.01^{\phantom{***}}$ & $0.49$ &  & $0.00^{\phantom{***}}$ & $0.52$ &  & $0.00^{\phantom{***}}$ & $0.53$ & \\
8 & $0.03^{*\phantom{**}}$ & $0.00$ & ${}^{-+-}$ & $0.02^{**\phantom{*}}$ & $0.49$ & ${}^{-+-}$ & $0.02^{*\phantom{**}}$ & $0.53$ & ${}^{-+-}$ & $0.02^{*\phantom{**}}$ & $0.53$ & ${}^{-+-}$\\
9 & $0.03^{*\phantom{**}}$ & $0.00$ & ${}^{-+-}$ & $0.02^{*\phantom{**}}$ & $0.49$ & ${}^{-+-}$ & $0.02^{*\phantom{**}}$ & $0.53$ & ${}^{-+-}$ & $0.02^{*\phantom{**}}$ & $0.53$ & ${}^{-+-}$\\
10 & $0.02^{*\phantom{**}}$ & $0.00$ & ${}^{-+-}$ & $0.01^{\phantom{***}}$ & $0.49$ &  & $0.01^{\phantom{***}}$ & $0.53$ &  & $0.01^{\phantom{***}}$ & $0.53$ & \\
11 & $0.04^{**\phantom{*}}$ & $0.01$ & ${}^{++-}$ & $0.01^{\phantom{***}}$ & $0.49$ &  & $0.01^{\phantom{***}}$ & $0.53$ &  & $0.01^{\phantom{***}}$ & $0.53$ & \\
12 & $0.11^{***}$ & $0.02$ & ${}^{++-}$ & $0.01^{\phantom{***}}$ & $0.49$ &  & $0.01^{\phantom{***}}$ & $0.53$ &  & $0.01^{\phantom{***}}$ & $0.53$ & \\
13 & $0.04^{**\phantom{*}}$ & $0.01$ & ${}^{++-}$ & $0.01^{\phantom{***}}$ & $0.49$ &  & $0.01^{\phantom{***}}$ & $0.53$ &  & $0.01^{\phantom{***}}$ & $0.53$ & \\
14 & $0.07^{***}$ & $0.01$ & ${}^{++-}$ & $0.02^{*\phantom{**}}$ & $0.49$ & ${}^{-++}$ & $0.00^{\phantom{***}}$ & $0.52$ &  & $0.00^{\phantom{***}}$ & $0.53$ & \\
15 & $0.04^{**\phantom{*}}$ & $0.01$ & ${}^{++-}$ & $0.00^{\phantom{***}}$ & $0.49$ &  & $0.00^{\phantom{***}}$ & $0.52$ &  & $0.00^{\phantom{***}}$ & $0.53$ & \\
16 & $1.03^{***}$ & $0.14$ & ${}^{+++}$ & $0.06^{***}$ & $0.49$ & ${}^{+++}$ & $0.04^{***}$ & $0.53$ & ${}^{-++}$ & $0.04^{***}$ & $0.53$ & ${}^{-++}$\\
17 & $1.33^{***}$ & $0.18$ & ${}^{+++}$ & $0.09^{***}$ & $0.50$ & ${}^{+++}$ & $0.07^{***}$ & $0.53$ & ${}^{-++}$ & $0.07^{***}$ & $0.54$ & ${}^{-++}$\\
18 & $0.04^{**\phantom{*}}$ & $0.01$ & ${}^{++-}$ & $0.01^{*\phantom{**}}$ & $0.49$ & ${}^{-++}$ & $0.00^{\phantom{***}}$ & $0.52$ &  & $0.00^{\phantom{***}}$ & $0.53$ & \\
19 & $0.06^{**\phantom{*}}$ & $0.01$ & ${}^{++-}$ & $0.00^{\phantom{***}}$ & $0.49$ &  & $0.00^{\phantom{***}}$ & $0.52$ &  & $0.00^{\phantom{***}}$ & $0.53$ & \\
20 & $0.16^{***}$ & $0.02$ & ${}^{++-}$ & $0.00^{\phantom{***}}$ & $0.49$ &  & $0.00^{\phantom{***}}$ & $0.52$ &  & $0.00^{\phantom{***}}$ & $0.53$ & \\
21 & $0.22^{***}$ & $0.03$ & ${}^{+++}$ & $0.01^{\phantom{***}}$ & $0.49$ &  & $0.01^{\phantom{***}}$ & $0.52$ &  & $0.00^{\phantom{***}}$ & $0.53$ & \\
22 & $0.58^{***}$ & $0.08$ & ${}^{+++}$ & $0.01^{\phantom{***}}$ & $0.49$ &  & $0.04^{***}$ & $0.53$ & ${}^{+++}$ & $0.04^{***}$ & $0.53$ & ${}^{+++}$\\
23 & $0.29^{***}$ & $0.04$ & ${}^{+++}$ & $0.00^{\phantom{***}}$ & $0.49$ &  & $0.00^{\phantom{***}}$ & $0.52$ &  & $0.00^{\phantom{***}}$ & $0.53$ & \\
    \hline
    \end{tabular}
    \caption{Results of reading time modeling with Natural Stories and Mamba-130M. $\Delta$MSE indicates the performance of $\bar\Delta_t$. The three signs in the $\beta$ rows indicate the direction of the effect of $\bar\Delta_t$ on the $i$th word due to the $\Delta_t$ value of the $i$th word, the $(i-1)$th word, and the $(i-2)$th word, respectively. $\Delta$MSE is calculated using cross validation, while $R^2$ and $\beta$ are calculated by fitting a single regression model to the entire data. Significance code: ${}^{***}:p<0.001, {}^{**}:p<0.01; {}^*:p<0.05$}\label{tab:modeling-results-natstor-130m}
\end{table*}

\begin{table*}[t]
    \small
    \centering
    \begin{tabular}{lll|rrr|rrr|rrr|rrr}
        \hline
        & & & & & & & & & \multicolumn{3}{c|}{Low-level +} & \multicolumn{3}{c}{Low-level +}\\
        & \multicolumn{2}{l|}{Baseline$\to$} & \multicolumn{3}{c|}{Intercept-only} & \multicolumn{3}{c|}{Low-level variables} & \multicolumn{3}{c|}{Mamba surprisal} & \multicolumn{3}{c}{GPT-2 surprisal}\\
        Model & \multicolumn{2}{l|}{$\downarrow$Data} & \#${}_\text{sig}$ & \#${}_\text{ind}$ & $R^2_\text{Max}$ & \#${}_\text{sig}$ & \#${}_\text{ind}$ & $R^2_\text{Max}$ & \#${}_\text{sig}$ & \#${}_\text{ind}$ & $R^2_\text{Max}$ & \#${}_\text{sig}$ & \#${}_\text{ind}$ & $R^2_\text{Max}$\\
        \hline
        \multirow{3}{1.5cm}{130M\\(24 layers)} & NS & SPR & $20$ & $16$ & $0.18_{17}$ & $10$ & $5$ & $0.50_{17}$ & $8$ & $6$ & $0.53_{0\phantom{0}}$ & $8$ & $6$ & $0.54_{17}$\\
        \cline{2-15}
        & \multirow{2}{*}{OS} & FP & $24$ & $19$ & $0.04_{7\phantom{0}}$ & $23$ & $11$ & $0.19_{7\phantom{0}}$ & $24$ & $11$ & $0.20_{13}$ & $21$ & $5$ & $0.21_{13}$\\
        & & RP & $20$ & $16$ & $0.01_{23}$ & $11$ & $5$ & $0.08_{21}$ & $16$ & $7$ & $0.09_{21}$ & $20$ & $5$ & $0.10_{11}$\\
        \hline
        \multirow{3}{1.5cm}{2.8B\\(64 layers)} & NS & SPR & $64$ & $25$ & $0.22_{41}$ & $29$ & $21$ & $0.51_{41}$ & $32$ & $22$ & $0.53_{41}$ & $26$ & $20$ & $0.54_{41}$\\
        \cline{2-15}
        & \multirow{2}{*}{OS}  &FP & $64$ & $33$ & $0.03_{2\phantom{0}}$ & $64$ & $19$ & $0.19_{18}$ & $56$ & $14$ & $0.20_{40}$ & $61$ & $13$ & $0.21_{42}$\\
        & & RP & $46$ & $27$ & $0.01_{60}$ & $28$ & $13$ & $0.09_{47}$ & $38$ & $13$ & $0.09_{47}$ & $42$ & $13$ & $0.10_{47}$\\
        \hline
    \end{tabular}
    \caption{Results of reading time modeling across datasets and model sizes. \#${}_\text{sig}$ indicates the number of layers whose predictive power is significant. \#${}_\text{ind}$ indicates the number of layers that are found to be independently predictive in the best model search. $R^2_\text{Max}$ is the value of the regression model with the best-performing layer, with the subscript indicating the index of the best-performing layer. NS = Natural Stories; OS = OneStop; SPR = self-paced reading; FP = first pass time; RP = regression path duration.}\label{tab:modeling-results-others}
\end{table*}

\subsection{Results}
The results with Natural Stories and Mamba-130M are shown in detail in Table \ref{tab:modeling-results-natstor-130m}. The results across datasets and models are summarized in Table \ref{tab:modeling-results-others}. The general tendencies are similar across datasets and models. We find that $\bar\Delta_t$ from most layers are predictive of human reading times when it is the sole predictor in the regression model. The correlations between $\bar\Delta_t$ and human reading times are not generally high, but reach as high as $R^2=0.22$ (layer 41 of Mamba-2.8B, Natural Stories; cf. $R^2=0.21$ for GPT-2 surprisal in the same dataset). Across datasets and models, predictive power of some layers diminishes when control variables are taken into account, but some other layers remain significant. Also across datasets and models, multiple layers independently contribute to the prediction. Overall, these results indicate that $\bar\Delta_t$ is predictive of human reading times, and it explains a unique variance that is not explained by major predictors considered in the literature.

\subsection{Follow-up analyses}
% $\Delta_t$の読み時間に対する予測力は何に由来するのかを理解するため，$\Delta_t$についてさらなる分析を行った．
We conducted post-hoc analyses to understand what makes the discretization timestep predictive of human reading times. Below, we focus on reading times from Natural Stories Corpus and $\Delta_t$ from Mamba-130M.

% \subsection{言語学的特徴にみる層間の分担}
\subsubsection{Layers' sensitivity to linguistic features}
\label{sec:linguistic_analysis}

\begin{figure}[t]
\centering
  \includegraphics[width=\linewidth]{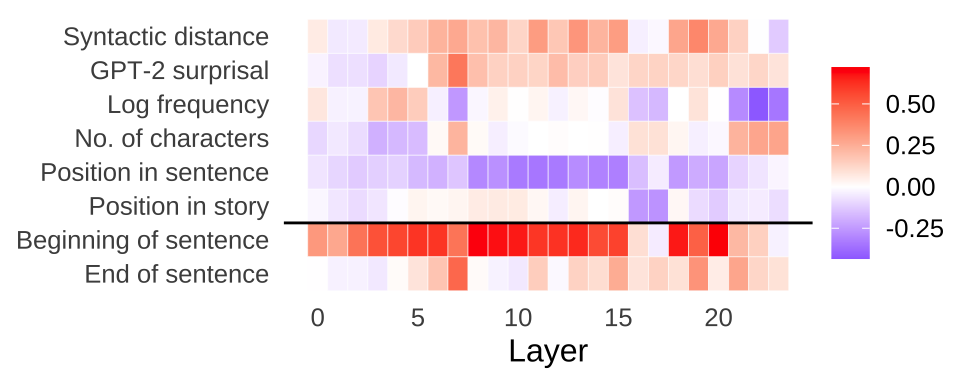}
  \caption{Correlations between linguistic features of $w_t$ and the discretization timestep $\bar\Delta_t$. Correlations for the beginning and end of sentence flags are calculated using all words in the corpus; others are calculated only using words that are not at the beginning or at the end of the sentence.}\label{fig:correlation}
  % \caption{単語$w_t$の言語学的な特徴と$\Delta_t$の相関．「文頭」「文末」については全単語で算出，それ以外については文頭・文末のトークンを除いて算出（読み時間分析に従う）．}
\end{figure}

\begin{figure}[t]
    \centering
    \includegraphics[width=0.7\linewidth]{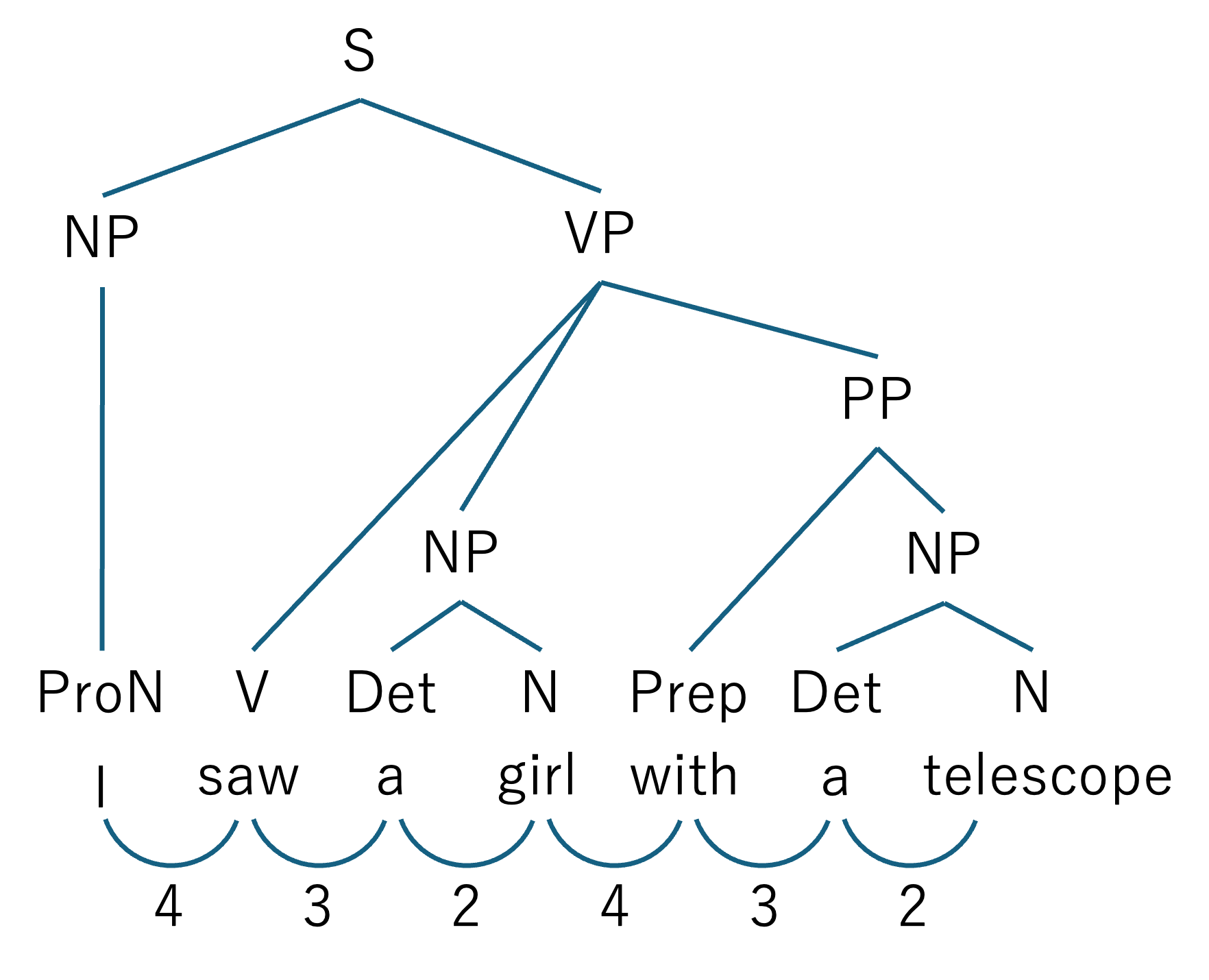}
    % \caption{構文木とそのパス長の例．大きな構造が現れるときのパスは長くなる傾向がある．}
    \caption{An example syntactic tree and the syntactic distance between adjacent words.}
    \label{fig:example_tree}
\end{figure}

Correlations between linguistic features of $w_t$ in the dataset and the discretization timestep $\bar\Delta_t$ are shown in Figure \ref{fig:correlation}. Linguistic features include those used in the reading time regression, in addition to boolean flags indicating whether the word is at the beginning or at the end of the sentence, and the distance between the previous and the current words on the Penn TreeBank--style syntactic tree (Figure \ref{fig:example_tree}).
% $w_t$の言語学的な諸特徴と$\Delta_t$との相関を図\ref{fig:correlation}に示す．諸特徴としては，読み時間の回帰に用いたもののほか，文頭・文末（該当すれば1，しなければ0），および前単語からの統語木上のパス長を含めた．統語木上のパス長は，単語同士の文法構造上の距離を示す（補遺\ref{appendix:path-length}参照）．

Two consistent trends are observed across layers: negative correlations between $\bar\Delta_t$ and the position of the word in the sentence, and positive correlations between $\bar\Delta_t$ and the beginning of the sentence. These correlations suggest that many layers undergo relatively large changes in their state at the beginning of the sentence, while they tend to maintain information within a sentence. Most layers also showed positive correlations between $\bar\Delta_t$ and syntactic distance. That is, $\bar\Delta_t$ of these layers increase when a large syntactic constituent completes and/or a new one begins (see Figure \ref{fig:example_tree}). These correlations suggest that the timestep sizes in the evolution of states in Mamba are sensitive to basic linguistic features.

% 全層でほぼ一貫した傾向としては，文中の位置（何単語目か）と$\Delta_t$が負に相関し，文頭と$\Delta_t$は正に相関した．これは，多くの層が文頭でその保持する情報を大きく変化させる一方，文の内部では情報を保持する傾向にあることを示す．また，統語木のパス長とは正の相関がみられた．これは，構造上の切れ目（たとえば，主語と動詞の間や，\textit{and}で結ばれた2つの節の間）では層の持つ情報が大きく変化することを示す．いずれも，Mamba の $\Delta_t$ が言語の基本的な構造を捉えていることを示唆する．

There are also distinct patterns found in particular layers, suggesting some division of labor between layers. Layers 16 and 17 clearly differ from other layers in the linguistic features they are sensitive to. $\bar\Delta_t$ from these layers negatively correlate with the position of the word in the story, while their correlation with the position of the word in the sentence, the beginning of the sentence, and the syntactic distances are weaker than the surrounding layers. This means that these layers are not very sensitive to local information within a sentence, but rather maintain story-level information. The particularly high predictive power that these layers showed in the reading time modeling (see Table \ref{tab:modeling-results-natstor-130m}) suggests that there may be some parallel between the way these layers maintain story-level information and the way humans do.

% 層間での分業もみられる．16・17層は周辺の層とはっきりと違った特徴を見せる．物語中の位置と負の相関がある一方，文中位置・文頭・統語木パス長との相関が弱い．これは，これらの層が文内部の局所的な構造にあまり反応せず，むしろ物語全体の情報を保持していることを示唆する．また，対数頻度とは負の相関があり，これは低頻度語に関する情報をよく取り込むことを意味する．一般には低頻度な語であっても，特定の物語ではよく現れることがある（本コーパスに含まれる物語では，主人公の名前\textit{Abby}や，主題の\textit{alien}など）ため，物語全体の理解にはこうした語の情報は有効である．こうした特徴を持つ16・17層が，読み時間モデリングで際立って高い予測力を示したことは，これらの層の振る舞いが人間の物語読解における長い文脈の保持のあり方を捉えていることを示唆する．この他，7層や21〜23層も特徴的であるが，紙幅の都合もあり詳細な検討は割愛する．

% 7層、21〜23層・・・？

\subsubsection{Analysis of Long-range Dependencies through Interventions
% in the Passkey Retrieval Task
}
\label{sec:passkey}

\begin{figure}[t]
  \centering
  \includegraphics[width=\linewidth]{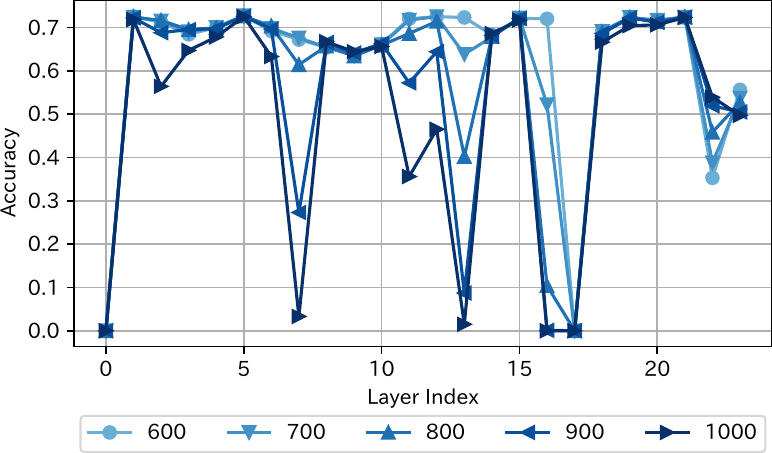}
  \caption{Accuracy in the passkey retrieval task when individually knocking out the SSM of each layer. The legend indicates the maximum token length of the input.}
  \label{fig:knockout}
\end{figure}

As shown in Table \ref{tab:modeling-results-natstor-130m}, layers 16 and 17 exhibit particularly high predictive power for Natural Stories reading time. 
These layers show distinct sensitivity to linguistic features, suggesting a role in maintaining story-level information (Figure \ref{fig:correlation}). 
To examine whether layers 16 and 17 actually capture long-range dependencies, we evaluate the contribution of each layer on the passkey retrieval task \cite{NEURIPS2023_ab05dc8b} using an intervention-based analysis.
The passkey retrieval task is commonly used to evaluate a model's ability to handle long-range dependencies. 
It assesses whether a model can correctly recover a specific string from an input consisting of that string embedded within a sequence of noise tokens.

\paragraph{Settings}
The experimental setup is as follows.
The model is given an input that begins with a six-digit number, followed by a sequence of noise sentences, and finally a prompt instructing it to reproduce the passkey. 
Concretely, the input takes the following form: ``The passkey is 317451. The grass is green. The sky is blue. The sun is yellow. Here we go. There and back again. ... The passkey is''. 
The output is considered correct if it exactly matches the passkey provided at the beginning of the input. 
The distance between the passkey and the output position is controlled by repeating the noise sentences until a specified length is reached. 
In this experiment, the length of the noise was adjusted so that the total input length was approximately \{600, 700, 800, 900, 1000\} tokens.
% \footnote{The total length of the input settings is not always divisible by the number of tokens in the noise sentence, so the noise sentence was repeated up to the maximum length that fits within the specified length.}
For each length setting, 1,000 input instances are generated, and accuracy is evaluated accordingly.
We measured the performance of the Mamba models with one layer knocked out at a time,
by setting the coefficient $\overline{B}_t$ for the input in Equation \eqref{eq:hidden-states} to zero at all times, thereby ensuring that $\bm{h}_t = 0$ is always maintained.

\paragraph{Results}
The results are shown in Figure \ref{fig:knockout}.
As illustrated, a notable drop in accuracy is observed when layer 0, 7, 13, 16, or 17 is knocked out. 
In particular, even when the input length is reduced to 800 tokens, layer 16 maintains an accuracy in the 10\% range, and for layers 0 and 17, the accuracy remains at 0\% even when the input is shortened to 600 tokens. 

\paragraph{Discussions}
The results suggest some layers are essential for inference that requires the transmission of information over long-range context.
However, it should be noted that these layers are not necessarily specialized for long-range dependencies; it remains possible that they are simply essential for text processing in general.
Interestingly, the layers 16 and 17 of Mamba-130M are not particularly predictive of OneStop reading time, unlike in the case of Natural Stories. One possible interpretation of this discrepancy is that human readers read differently depending on the reading paradigm and/or materials. For example, the self-paced reading paradigm might encourage readers to spend more time when the input word is associated with a distant context since the retrieval of context relies solely on memory, while in the eyetracking paradigm the retrieval can be done by physically looking at the context. This is a pure speculation, and we leave it for future work exactly why different layers are responsible for predicting reading times from different datasets.

\section{Mamba as a New Lens on Human Language Processing}\label{sec:lens}
% \label{sec:layer16-17-analysis}
% \textcolor{red}
{Having established an empirical connection between the discretization timestep $\bar\Delta_t$ and human reading times, we turn to theoretical discussions on how Mamba processes language over time with ever-updated memory. We suggest that the model can be informative to the science of language processing by humans, which also process language in an incremental fashion under memory constraints.}

\subsection{Transition Matrix Eigenspectrum as Memory Retention Ability}
\label{sec:eigen}

The extent to which Mamba attenuates memory of past context depends on the elements of the transition matrix $\exp(A)$. 
In this section, we first explain this property analytically and then verify that layers 16 and 17, which were effective for predicting reading time, also exhibit distinctive behavior in terms of the eigenvalues of the transition matrix.

Since the transition matrix $\exp(A)$ of Mamba is diagonal, its eigenvalues are simply its diagonal elements.
In Equation \eqref{eq:hidden-states}, when the elements of the transition matrix are close to one, that is, when $\exp(A)\approx I$, we have $\overline{A}_t \approx I$ regardless of the value of $\Delta_t$, and thus the past context $\bm{h}_{t-1}$ is retained over the long term.
In contrast, when the elements of the transition matrix are close to zero, that is, when $\exp(A)\approx O$, we have $\overline{A}_t \approx O$, and the past context $\bm{h}_{t-1}$ is rapidly forgotten.
In this way, the magnitude of the eigenvalues of the transition matrix $\exp(A)$ serves as an indicator for evaluating the ability of each SSM layer to retain information over long timescales, from a perspective distinct from the discretization timestep $\Delta_t$.

% The long-term information retention capability of each layer in a state-space model can be evaluated by the eigenvalues of the transition matrix $\exp(A)$.
% The absolute value of an eigenvalue being closer to 1 indicates information is more persistently stored, while being closer to 0 indicates forgetting.

\begin{figure}[t]
  \centering
  \includegraphics[width=\linewidth]{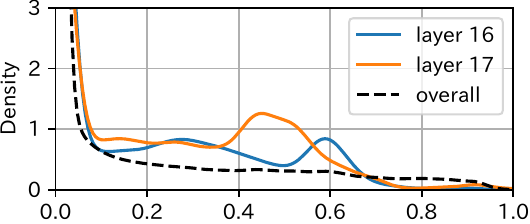}
  \caption{Eigenvalue distribution of the transition matrix $\exp(A)$. 
  This plots the eigenvalue distribution over all layers and separately for layers 16 and 17, where timestep $\Delta_t$ does not react to sentence boundaries.}
  \label{fig:eigen_kde_plot}
\end{figure}

Figure \ref{fig:eigen_kde_plot} shows the eigenvalue distributions for the entire layer and for layers 16 and 17, where timestep $\Delta_t$ does not react to the sentence boundary.
From the figure, it can be seen that while the overall eigenvalue distribution peaks near zero, the eigenvalue distributions of layers 16 and 17 show a higher proportion of eigenvalues near 0.5 compared to the overall distribution.
The eigenvalue analysis of layers 16 and 17 from the perspective of SSM dynamics thus agrees with the analysis based on linguistic features and $\Delta_t$ in suggesting that these layers tend to retain information over the long term.

% \textcolor{red}
{In short, layers of Mamba process information at different timescales via different eigenvalues of transition matrices (as well as varying $\Delta_t$), which can be interpreted as \emph{memory retention ability} of each layer. Interesting future directions based on this observation include comparing this mechanism to human brain activities, which also seem to keep track of information that unfolds at different timescales \cite{gwilliams-etal-2024-hierarchical}, with potential division of labor between neural populations \citep{gwilliams-etal-2022-neural}.}

% \subsection{Exploring the Relationship Between the Discretization Timestep $\Delta_t$ and State Transition Uncertainty}
\subsection{Discretization Timestep as State Transition Uncertainty}
\label{sec:uncertainty}

We theoretically show that the component $\Delta_t$ corresponds, in a sense, to the \emph{uncertainty} in state transitions.
% , in addition to representing the discretization timestep.
% The timestep $\Delta_t$ numerically represents the time increment used to discretize a continuous-time model (Figure \ref{fig:image}). 
% It suggests a concept analogous to human reading time, which tends to increase when prediction becomes difficult and uncertainty is high.

% ここから先の構成の suggest by 横井

% 以下 formal な statement を述べる。
% まず、mamba の隠れ状態の遷移を確率的モデルとして解釈するために、形式的にノイズ項を加える。
% つまり、元々の遷移の関数は式 \eqref{eq:hidden-states} であったが、これに微小な Gaussian noise を加える：
We now present the following formal statement. 
First, to interpret the hidden-state transitions of Mamba as a probabilistic model, we formally introduce a noise term. 
Specifically, while the original transition function is given by Equation \eqref{eq:hidden-states}, we add a small Gaussian noise term $w(t)$ to it as follows:
\begin{align}
  h'(t) &= A(t) h(t) + B(t) x(t) + w(t)\\
  w(t) &\sim \mathcal{N}(0,Q_c).
\end{align}
% こうすると、隠れ状態の遷移 $p(h_t\mid h_{t-1})$ を closed form で書けるようになる:
% （see Appendix hoge for more details）。
This allows the hidden-state transition $p(h_t \mid h_{t-1})$ to be written in closed form:
\begin{align}
  p(h_t \mid h_{t-1}) := \mathcal{N}\left(h_t ~;~ \overline{A}_t h_{t-1} + \overline{B}_t x_t, Q_d\right),
\end{align}
% ただし $Q_d$ は離散化後のノイズ強度である。
where $Q_d$ denotes the noise intensity after discretization.
% この標準的な仮定の下で、\emph{隠れ状態 h の遷移に関する条件付きエントロピー（不確実さ）が、Δt で決まる（Δt が増えれば不確実さが増える）}、ことがわかる：
Under this standard assumption, it follows that \emph{the conditional entropy (uncertainty) of the hidden-state transition $\bm{h}_t$ is determined by $\Delta_t$, with the uncertainty increasing as $\Delta_t$ becomes larger}:
\begin{align}
  H[h_{i+1} \mid h_i]
= \frac{1}{2}\sum_{i=1}^n \log\left(e^{2A_{ii}\Delta_t}-1\right) + \mathrm{const.},
\label{eq:uncertainty}
\end{align}
% ただし $A_{ii}$ は離散化前の遷移行列 $A$ の第$i$対角成分。
where $A_{ii}$ denotes the $i$-th diagonal element of the transition matrix $A$ before discretization.
% あとで元気（時間）があったら、「仮定軽いよ！安心してね！」という意図が伝わるような何かを書く
% 詳細な導出については Appendix \S\ref{apdx:uncertainty-derivation} を参照されたい。
For a detailed derivation, refer to Appendix \S\ref{apdx:uncertainty-derivation}.

% (以下、in practice の段落)
% First, we introduce uncertainty into Mamba's state transitions.
% Although Mamba's hidden state update rule is deterministic and assumes no noise, we consider a hypothetical scenario in which a noise term is added to illustrate how uncertainty after a state transition depends on the discretization timestep $\Delta_t$.

In practice, how does Mamba adjust the timestep $\Delta_t$ in response to the given input text?
Figure \ref{fig:delta_t_vis} shows that in several layers, the timestep $\Delta_t$ peaks at the beginning of sentence tokens (this trend is also quantitatively confirmed in the analysis in \S\ref{sec:linguistic_analysis} (Figure \ref{fig:correlation})).
Mamba increases timestep $\Delta_t$ at the beginning of sentences, precisely at the points where it is difficult to predict from the immediate preceding context (typically ending with periods).
As confirmed in this section, it is intuitive that a larger timestep $\Delta_t$ corresponds to greater uncertainty in state transition.
However, it is not trivial that Mamba possesses the property of incrementing $\Delta_t$ at uncertain points during state transition.
Why and how this property was acquired through learning remains an open question.
Finally, it should be noted that the considerations here are based on the interpretation when a noise term is introduced into Mamba and do not naturally follow from the original deterministic model.

% 本研究において，離散化ステップ $\Delta_t$ が読み時間モデリングに寄与した理由は，この性質がヒトの認知負荷の増減の特徴と対応しているためであると考えられる．

% \textcolor{red}
{It should be noted that entropy/uncertainty discussed here is \emph{not} directly related to the uncertainty about subsequent input much discussed and investigated in the cognitive modeling literature \citep{hale-2016-information}. Instead, we are looking at uncertainty about the state that model arrive at when the processing of the current word is complete. This notion of uncertainty may offer a new perspective on the influence of noise on language processing, a hotly discussed topic in computational psycholinguistics. In existing models of noisy language processing \cite{futrell-etal-2020-lossy, hahn-etal-2022-resource}, the noise erases individual words at a certain probability when each word is read. In the current assumption, noise is added to the spatial representation of the context and accumulates over time. Future work can explore whether such a view on noise better capture limitations and rationality of human language processing.}

\section{Related Works}

\citet{10.1162/TACL.a.58} performs cognitive modeling using internal components of language models, although it focuses on Transformer-based models rather than SSM-based ones. 
The authors demonstrate an intriguing correspondence between the ordering of layers and the temporal order of human responses: 
earlier layers align with early eye-movement measures, while later layers correspond to later responses associated with deeper semantic processing.
Their approach converts intermediate layer representations into values corresponding to surprisal and uses them for cognitive modeling.
In contrast, our study directly incorporates the internal model component $\Delta_t$, interpretable as the processing time of an SSM-based language model, into reading-time modeling, which constitutes a key difference from their approach.

\citet{sharma2024locating} and \citet{endy-etal-2025-mamba} propose methodologies for identifying layers that are responsible for transmitting factual information. 
Specifically, given the input ``Michael Jordan professionally played,'' they examine how the factual knowledge ``basketball'' is propagated through the model to the final output token. 
Through intervention experiments that disrupt information transmission pathways, both studies demonstrate that factual information is conveyed from the subject to the final token primarily via later intermediate layers.
These approaches appear promising as follow-up analyses for reading-time modeling, as they offer ways to probe what computations are performed in individual layers. 
However, their analyses focus on short factual sentences and do not address the processing of story-level long-form text as considered in this work; extending such analyses to longer narratives remains an open problem for the field.

Why, in the first place, do the discretization timesteps $\Delta_t$ peak at sentence boundaries?
\citet{liu-ding-2025-information} provides insights that help address this question.
They report that large language models, when processing information, constrain the scope of information integration at sentence boundaries, similarly to humans.
Their experiments focus on Qwen and Pythia models with attention mechanisms, which can attend to the entire input text.
Nevertheless, they show that these models tend to preferentially rely on words within the same sentence or clause.
The phenomenon observed in Figures \ref{fig:delta_t_vis} and \ref{fig:correlation} of our study, where peaks of timesteps $\Delta_t$ arise at sentence boundaries, likewise induces an effect that limits the range of information integration at such boundaries, consistent with their findings.
This is because timestep $\Delta_t$ functions as a gate: when it becomes large, it forgets past memory $h_{t-1}$ and incorporates more of the current input $x_t$.
However, layer 17 of Mamba-130M, which showed the highest predictive performance for Natural Stories in the current experiment, does not exhibit peaks at sentence boundaries.
This suggests that information transmission across sentences is an essential component explaining reading behavior, and thus our result differs in this respect from theirs.

% \begin{itemize}
%     \item echo state network での読み時間モデリング？: Insensitivity of the Human Sentence-Processing System to Hierarchical Structure（open access じゃないので読めない）
% \end{itemize}

\section{Conclusion}

% We showed that Mamba's discretization timestep $\Delta_t$ from multiple layers is predictive of human reading time.
% Our post-hoc analysis suggests that multiple processing mechanisms with distinct roles exist within Mamba, and that they contribute independently.
% This result implies that $\Delta_t$ represents a ``processing time'' that varies according to input, potentially reflecting cognitive load in part.

% This research clearly diverges from conventional reading time modeling in that we directly employ the discretization timestep $\Delta_t$ of the SSM (which corresponds to the gate in recurrent neural networks) for prediction.
% We expect this work to provide a completely novel perspective on reading time modeling theory and to contribute toward improving the interpretability of linear attention-based models, which are currently gaining significant attention.

We showed that Mamba's discretization timestep $\Delta_t$ is predictive of human per-word reading times. The predictive power of discretization timestep $\Delta_t$ from the best-performing layer is comparable to that of GPT-2 surprisal, and $\Delta_t$ from many layers remain significant even when GPT-2 surprisal and other known predictors are controlled for.
We also conducted formal analyses of Mamba's architecture and internal dynamics, and suggested that Mamba can serve as a new, valuable lens to look at human real-time language processing with ever-updated memory.
With this new lens, future work may investigate how humans, who are constrained by the flow of time and ever-changing memory, still show remarkable ability to use language that unfolds over time, often quite rapidly.

% Post-hoc analyses indicate that Mamba contains multiple processing mechanisms with distinct and independent roles, suggesting that $\Delta_t$ functions as an input-dependent ``processing time'' that partly reflects cognitive load. 
% Unlike conventional reading-time modeling, our approach directly uses the SSM discretization timestep $\Delta_t$, corresponding to a gate in recurrent neural networks. 
% We expect this work offers a novel perspective on reading-time modeling and contributes to the interpretability of linear attention-based models.

% \newpage
\section{Limitations}

We have not clarified how the model learns to increase discretization timestep $\Delta_t$ at sentence boundaries or when the syntactic path length to adjacent words is long.
Therefore, we report only that we discovered and examined that the timestep $\Delta_t$ indeed contributes to the reading time modeling.
To understand how the timestep $\Delta_t$ peaks at sentence boundaries and other locations, statistical analysis of the training corpus and analysis of the learning dynamics would be necessary, but this is beyond the scope of this paper.

% Our experiments showed that the discretization timesteps $\Delta_t$ across multiple layers contributed to read time modeling.
% However, the consideration of why each value contributed to the modeling is suggestive and has not been clarified through rigorous examination via intervention experiments or similar methods.
The analysis in Section \S\ref{sec:passkey} provided evidence that layer 17 of Mamba-130M, which showed the strongest contribution to predicting Natural Stories reading time, potentially captures long-range dependencies.
However, this does not directly imply that long-range dependencies are inherently a crucial factor for modeling reading time.
In other words, a principled explanation for why the timestep $\Delta_t$ contributes to reading time modeling has yet to be established.

Since experiments were conducted solely using Mamba, the discussions in this paper cannot be applied to other architectures.
The reason only one architecture was adopted is that Mamba is the only one where the model itself dynamically determines the discretization timestep based on SSM.

\section*{Acknowledgments}

This work was supported by JST SPRING (JPMJSP2104), JST FOREST (JPMJFR2331), and JSPS KAKENHI (JP22H05106).

This study was inspired by discussions held at the weekly ``317 Café'' at the National Institute for Japanese Language and Linguistics (NINJAL).\footnote{\url{https://sites.google.com/view/317cafe/}} We would like to express our sincere gratitude to all participants of the 317 Café.

\bibliography{custom}

\appendix

\onecolumn
\section{Derivation of State Transition Uncertainty}
\label{apdx:uncertainty-derivation}

In this section, we derive Equation \eqref{eq:uncertainty}.
The state equations for a continuous-time SSM with a Gaussian noise term $w(t)$ are as follows:
\begin{align}
  h'(t) &= A(t) h(t) + B(t) x(t) + w(t)\\
  w(t) &\sim \mathcal{N}(0,Q_c),
\end{align}
where $Q_c$ denotes the noise intensity in continuous time and is assumed to be diagonal, similar to the transition matrix $A$ of Mamba.
% これは，状態空間モデルを次のような確率モデルとみなしていることに他ならない：
This is equivalent to interpreting the state-space model as the following probabilistic model:
\begin{align}
  p(h_t \mid h_{t-1}) := \mathcal{N}\left(h_t ~;~ \overline{A}_t h_{t-1} + \overline{B}_t x_t, Q_d\right),
  \label{eq:ssm-prob-model}
\end{align}
% 右辺は，平均 $\overline{A}_t h_{t-1} + \overline{B}_t x_t$，分散 $Q_d$ の多変量正規分布を表す．
where $\overline{A}_t$ and $\overline{B}_t$ denote the SSM parameters after discretization (Eq. \eqref{eq:ssm-param1} and \eqref{eq:ssm-param2}), and $Q_d$ denotes the noise intensity after discretization.

Applying Zero-Order Hold (ZOH) discretization yields the following noise term:
\begin{align}
  w_t \sim \mathcal{N}(0, Q_d), ~~
  Q_d = \int_{0}^{\Delta_t} e^{A\tau} Q_c e^{A^\top\tau} d\tau,
  \label{eq:discrete_noise}
\end{align}
where $Q_d$ is the discrete representation of the continuous-time noise intensity $Q_c$ at the sampling interval $\Delta_t$, and is a diagonal matrix.
% \comment{引用が欲しいが wiki 以外見つからない．}
Since the transition matrix $e^A$ of Mamba is diagonal,
the integrand is also a diagonal matrix, allowing Equation \eqref{eq:discrete_noise} to be solved as follows:
\begin{align}
  (Q_d)_{ii} &= \int_{0}^{\Delta_t} e^{2A_{ii}\tau} (Q_c)_{ii} d\tau 
  = (Q_c)_{ii} \frac{e^{2A_{ii} \Delta_t}-1}{2A_{ii}}.
  \label{eq:discrete_noise_diagonal}
\end{align}

The uncertainty before and after the state transition can be quantified by the conditional differential entropy $H[h_t \mid h_{t-1}]$:
\begin{align}
  H[h_t \mid h_{t-1}] 
  = - \int_{\mathbb{R}^n} p(h_t \mid h_{t-1}) \log p(h_t \mid h_{t-1}) dh_t.
  \label{eq:conditional_entropy}
\end{align}
Here, by performing the following change of variables,
$\mathcal{N}(h_t ~;~ \overline{A}_t h_{t-1} + \overline{B}_t x_t, Q_d)$
is equivalent to $\mathcal{N}(w_t ~;~ 0, Q_d)$:
\begin{align}
  &\quad\mathcal{N}(h_t ~;~ \overline{A}_t h_{t-1} + \overline{B}_t x_t, Q_d) \\
  &= \frac{1}{\sqrt{(2\pi)^n \lvert Q_d \rvert}} 
    \exp\left(-\frac{1}{2}(h_t - (\overline{A}_t h_{t-1} + \overline{B}_t x_t))^\top Q_d^{-1} (h_t - (\overline{A}_t h_{t-1} + \overline{B}_t x_t))\right) \\
  &= \frac{1}{\sqrt{(2\pi)^n \lvert Q_d \rvert}} 
    \exp\left(-\frac{1}{2} w_t^\top Q_d^{-1} w_t\right) \\
  &= \mathcal{N}( w_t ~;~ 0, Q_d).
\end{align}
Then, Equation \eqref{eq:conditional_entropy} can be rewritten as follows from the above equation and Equation \eqref{eq:ssm-prob-model}:
\begin{align}
  H[h_t \mid h_{t-1}] 
  &= - \int_{\mathbb{R}^n} \mathcal{N}( w_t ~;~ 0, Q_d) \log \mathcal{N}( w_t ~;~ 0, Q_d) dw_t.
\end{align}
Since the right-hand side is the differential entropy of a Gaussian distribution, it is given by the following:
\begin{align}
  H[h_t \mid h_{t-1}]
  = \frac{1}{2}\log\left((2\pi e)^n \lvert Q_d \rvert\right)
\end{align}
Here, substituting the components of $Q_d$ derived earlier in Equation \eqref{eq:discrete_noise_diagonal}, we obtain the following expression:
\begin{align}
  H[h_{i+1} \mid h_i]
= \frac{1}{2}\sum_{i=1}^n \log\left(e^{2A_{ii}\Delta_t}-1\right) + \mathrm{const.}
\end{align}
where the terms independent of timestep $\Delta_t$ are grouped as a constant term.

Therefore, the uncertainty in state transitions monotonically increases with respect to the discretization timestep $\Delta_t$.
Conversely, as the timestep $\Delta_t$ approaches zero, the differential entropy tends to $-\infty$, which indicates the absence of uncertainty.
This leads to the intuitive conclusion: if the sampling period is coarse, the uncertainty regarding the next state increases; conversely, if sampled finely, the uncertainty becomes sufficiently small.

\end{document}